\pdfoutput=1
\documentclass{article}
\usepackage{microtype} 
\usepackage{graphicx}
\usepackage{float}
\usepackage{placeins}
\usepackage{balance}
\usepackage{etoolbox}
\makeatletter
\patchcmd{\@BAlancecol}{\setbox\@leftcolumn}{\advance\@tempdima by 3pt\setbox\@leftcolumn}{}{}
\makeatother
\usepackage{booktabs}
\usepackage{array,tabularx}
\newcolumntype{C}{>{\centering\arraybackslash}X}

\usepackage{amsmath,amssymb}
\usepackage{xcolor}
\usepackage{colortbl}
\definecolor{LIETableRow}{rgb}{0.90353,0.91373,0.94118}
\definecolor{LIEStaticRow}{HTML}{E7F0E5}
\definecolor{LIEDynamicRow}{HTML}{FBEDD9}
\usepackage{xurl}
\usepackage{hyperref}
\hypersetup{
  hypertexnames=false,
  colorlinks=true,
  linkcolor=blue!50!black,
  citecolor=blue!50!black,
  urlcolor=blue!50!black
}

\usepackage[accepted]{mlsys2025}
\makeatletter
\renewcommand{\Notice@String}{Preprint.}
\makeatother
\usepackage{algorithm}
\usepackage{algorithmic}
\usepackage{enumitem}
\usepackage{tikz}
\usetikzlibrary{arrows.meta,positioning}

\usepackage{tcolorbox}
\tcbuselibrary{skins,breakable}
\definecolor{LIEInk}{rgb}{0.25882,0.28432,0.35294}
\definecolor{LIEPaper}{rgb}{0.90353,0.91373,0.94118}
\tcbset{liecallout/.style={
  enhanced,
  breakable=false,
  width=\linewidth,
  colback=LIEPaper,
  colframe=black,
  colbacktitle=LIEInk,
  coltitle=white,
  fonttitle=\normalfont\normalsize\strut,
  halign title=center,
  fontupper=\small,
  boxrule=0.5mm,
  arc=1mm,
  outer arc=1.5mm,
  boxsep=0pt,
  left=8pt,right=8pt,top=10pt,bottom=8pt,
  before skip=12pt,after skip=9pt,
  attach boxed title to top left={xshift=8pt,yshift=-2mm},
  boxed title style={boxrule=0pt,arc=0.75mm,outer arc=0.75mm,
    colframe=LIEInk,colback=LIEInk,boxsep=0pt,halign=center,
    left=3mm,right=3mm,top=2.5pt,bottom=2.5pt},
  before upper={\setlength{\parindent}{0pt}\setlength{\parskip}{3pt}}
}}
\newtcolorbox{keybox}[1][Key Findings]{liecallout,title={#1}}
\newtcolorbox{researchquestion}[1][]{
  liecallout,title={Research Question},fontupper=\small\itshape,#1
}
\newtcolorbox{takeaway}[1][]{
  enhanced,breakable=false,width=\linewidth,
  colback=black!2,colframe=black!12,colbacktitle=black!12,coltitle=black,
  boxrule=0.3pt,titlerule=0pt,arc=2mm,outer arc=2mm,boxsep=0pt,
  left=8pt,right=8pt,top=6pt,bottom=6pt,toptitle=4pt,bottomtitle=4pt,
  before skip=7pt,after skip=7pt,
  fonttitle=\normalsize\bfseries,fontupper=\normalsize,
  before upper={\raggedright\setlength{\parindent}{0pt}\setlength{\parskip}{0pt}},#1
}
\newtcolorbox{questionanswer}{
  enhanced,breakable=false,width=\linewidth,
  colback=black!7,frame hidden,boxrule=0pt,
  arc=2mm,outer arc=2mm,boxsep=0pt,
  left=8pt,right=8pt,top=6pt,bottom=6pt,
  before skip=7pt,after skip=7pt,fontupper=\normalsize,
  before upper={\raggedright\setlength{\parindent}{0pt}\setlength{\parskip}{0pt}}
}
\newtcolorbox{keyfindings}[1][]{liecallout,title={Key Findings},#1}

\newcommand{\R}{\mathbb{R}}
\newcommand{\clip}{\operatorname{clip}}
\newcommand{\argmin}{\operatorname*{arg\,min}}

\newcommand{\method}[1]{\texttt{#1}}

\mlsystitlerunning{LEAP: Generalizable Length Prediction for LLM Scheduling}

\hypersetup{
  pdfauthor={Kwan Yan, Jialun Zhong, Yanzeng Li},
  pdfsubject={arXiv preprint}
}

\begin{document}
\raggedbottom
\renewcommand{\topfraction}{0.95}
\renewcommand{\bottomfraction}{0.9}
\renewcommand{\textfraction}{0.05}
\renewcommand{\floatpagefraction}{0.8}
\setcounter{topnumber}{3}
\setcounter{bottomnumber}{2}
\setcounter{totalnumber}{5}
\setlength{\parskip}{4pt}
\setlength{\abovedisplayskip}{5pt plus 1pt minus 1pt}
\setlength{\belowdisplayskip}{5pt plus 1pt minus 1pt}
\setlength{\belowdisplayshortskip}{3pt plus 1pt minus 1pt}
\setlength{\intextsep}{10pt plus 1pt minus 1pt}
\setlength{\textfloatsep}{12pt plus 2pt minus 2pt}
\setlength{\dbltextfloatsep}{14pt plus 2pt minus 2pt}
\clubpenalty=10000
\widowpenalty=10000
\displaywidowpenalty=10000
\brokenpenalty=10000 
\twocolumn[
\mlsystitle{LEAP: Efficient LLM Inference Scheduling\texorpdfstring{\\}{ }
with Generalizable Length Prediction}
\begin{center}
\setlength{\tabcolsep}{13pt}
\begin{tabular}{@{}ccc@{}}
\textbf{Kwan Yan} & \textbf{Jialun Zhong} & \textbf{Yanzeng Li}\textsuperscript{*} \\
{\small Beijing Normal University} & {\small Peking University} & {\small Beijing Normal University} \\
{\small\href{mailto:yankuan@mail.bnu.edu.cn}{\texttt{yankuan@mail.bnu.edu.cn}}} &
{\small\href{mailto:zhongjl@stu.pku.edu.cn}{\texttt{zhongjl@stu.pku.edu.cn}}} &
{\small\href{mailto:liyanzeng@bnu.edu.cn}{\texttt{liyanzeng@bnu.edu.cn}}}
\end{tabular}
\end{center}
\mlsyskeywords{LLM serving, length prediction, scheduling, linear regression, online adaptation}
\vskip 0.3in
\begin{abstract}
Length-aware scheduling reduces request latency in LLM serving by prioritizing shorter requests. However, despite their large model sizes, length predictors can suffer degraded ranking performance under domain and task shifts, and adapting them can require costly retraining. Their prediction overhead can further offset the benefits of scheduling.
To improve the generalization and efficiency of length-aware scheduling, we propose \textbf{LEAP}, a length prediction system with fast online adaptation and asynchronous CPU execution.
LEAP reuses the serving model's prefill hidden states, which contain information about the output length, and extracts this signal through a linear readout.
To adapt to substantial domain and task shifts, it updates the readout at test time through recursive least squares, using completed requests as supervision without gradient computation or historical-data replay.
Prediction and updates run asynchronously on CPU, while prediction and initial batch ordering overlap the remaining GPU prefill.
Across the evaluated cross-domain transfers, LEAP retains strong ranking performance, and online adaptation improves Spearman correlation by 64.7\% under a larger domain shift.
Serving experiments show that LEAP reduces average per-token latency by 15.9--40.8\% compared with the evaluated length predictors, while achieving 1.9\% higher throughput with $30{,}569\times$ fewer predictor parameters than the strongest evaluated baseline.
\end{abstract}
]
\begin{NoHyper}
\begingroup
\renewcommand{\thefootnote}{*}
\footnotetext[0]{Corresponding author.}
\endgroup
\end{NoHyper}

\section{Introduction}
\label{sec:intro}

\begin{figure*}[!t]
\centering
\includegraphics[width=\textwidth]{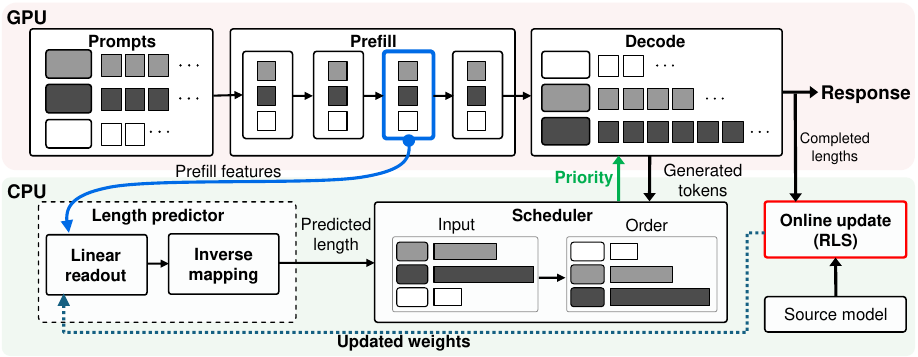}
\par\vspace{-6pt} 
\caption{\method{LEAP}: CPU-asynchronous prediction, scheduling, and completion-driven adaptation.}
\label{fig:architecture}
\vspace{-8pt} 
\end{figure*}

Length-aware scheduling plays a key role in LLM serving: it prevents long requests from blocking multiple short ones, reducing average response time~\citep{ssjf,ltr,trail,tie}. Its effectiveness depends on length predictions that distinguish requests by their remaining work. Yet deployment brings new tasks, prompt styles, and user populations. A predictor that performs well on its source domain may no longer provide useful priorities when these change.

This concern appears in our cross-domain evaluations (Section~\ref{sec:prediction-transfer}). On the same target requests, several predictors lose substantial ranking quality when trained on a different domain. Restoring that quality through repeated offline retraining adds data collection, optimization, and redeployment costs. Moreover, prediction itself also consumes resources: prompt-based methods run separate encoders~\citep{tie,ssjf,ltr}, while hidden-state methods reuse the serving model's representations through neural prediction heads~\citep{trail,star,egtp,estp}. A GPU-based predictor spends resources that scheduling is intended to use more effectively; a slow CPU predictor can leave the scheduler waiting. A practical design must therefore consider how well predictions transfer, how they can be updated, and what they cost to deliver.

For scheduling, \emph{sufficient prediction differs from perfect prediction.} Requests can be prioritized effectively even when their exact lengths remain uncertain. This motivates us to favor a lightweight predictor: higher accuracy does not improve serving efficiency when the additional prediction cost outweighs its scheduling benefits. We seek useful length information in features that the serving model already computes, avoiding the cost of a heavier predictor. Prior work uses hidden states for generation control and future-token prediction~\citep{houlatent,dflash}; hidden-state length predictors more directly demonstrate signals related to output length~\citep{trail,egtp}. This motivates reusing the representation produced during prefill, which is conditioned on both the prompt and the model that will generate the answer. We extract its length signal through a linear readout, whose small parameter set supports fast CPU inference and recursive updates.

To improve the generalization and efficiency of length-aware scheduling, we propose \method{LEAP}\footnote{Anonymous code repository: \url{https://anonymous.4open.science/r/LEAP-BBFF/}.}, a length prediction system with fast online adaptation and asynchronous CPU execution. Its readout uses the final prompt token's hidden state from one late layer to estimate output length for decode scheduling. When substantial domain or task shifts degrade the predictions, \method{LEAP} adapts the readout through test-time training with recursive least squares (RLS). Each completed request supplies its observed output length as supervision. Starting from the source-trained readout, RLS incorporates this feedback incrementally without backpropagation or replaying completed requests (Section~\ref{sec:adapt}).

What may be overlooked is the latency introduced by feature transfer, prediction, and score publication. This overhead can offset the benefits of more accurate predictions. To keep predictor computation off the GPU, \method{LEAP} runs prediction and updates asynchronously on CPU, allowing GPU decoding to continue using published scores. To reduce prediction delay, it overlaps CPU prediction and initial batch ordering with the remaining GPU prefill. In practice, both can finish before GPU prefill completes, as we verify in Section~\ref{par:prefill-overlap}.

Together, these components support decode scheduling as domains and tasks change. First, \method{LEAP} estimates output length from a single prefill hidden state. Second, it adapts the readout using completion feedback. Finally, it publishes predictions asynchronously for scheduling while GPU inference continues. Besides scheduling, our methodology may also benefit other use cases that involve output lengths, such as output length control through intervention and KV cache management.

Our contributions are as follows:
\begin{itemize}[topsep=4pt,partopsep=0pt,itemsep=3pt,parsep=0pt]
  \item \begin{samepage}\textbf{Reusing prefill signals for scheduling.} We reuse a single prefill hidden state to predict output length without a separate prompt encoder. The linear readout uses square-root target regression to balance tail compression and the separation of long requests (Section~\ref{sec:readout}).\par\end{samepage}
  \item \textbf{Adaptation through test-time training.} We adapt the source-trained readout to domain and task shifts using completion-supervised RLS updates, without gradient computation or historical-data replay (Section~\ref{sec:adapt}).
  \item \textbf{Asynchronous execution on CPU.} We integrate CPU prediction, online updates, and score publication into a vLLM serving path. Prediction and initial batch ordering overlap the remaining GPU prefill (Section~\ref{sec:system}).
  \item \textbf{Experimental results.} The frozen readout retains stronger ranking performance than the evaluated baselines across domain transfers. Online adaptation substantially improves ranking under larger language shifts. Serving experiments demonstrate measurable per-token latency improvements with significantly lower predictor parameter counts.
\end{itemize}

\section{Motivation: Rethinking Length Prediction for LLM Scheduling}
\label{sec:background}

\subsection{Q1: Does a length predictor generalize to new domains and tasks?}

A predictor trained on one dataset may encounter different tasks, prompt styles, and user populations after deployment. These shifts alter the relationship between prompts and output length, so a fixed-distribution test does not establish the robustness of its request rankings when this relationship changes.

Our cross-domain evaluation (Figure~\ref{fig:ood-bars}) shows that frozen predictors can lose substantial ranking quality under domain shifts. Recovering this quality through offline retraining is costly: it requires new data collection, model optimization, and redeployment. A practical design therefore needs two properties: baseline generalization to new domains, and an affordable mechanism to adapt when that generalization proves insufficient. Completed requests naturally provide feedback in the form of observed output lengths, offering a signal for adaptation without additional annotation effort.

\begin{takeaway}[title={Takeaway 1: Look beyond the source domain}]
Length predictors should retain useful rankings across domains and tasks and support affordable updates when those rankings degrade.
\end{takeaway}

\subsection{Q2: Does scheduling really need precise length predictions?}

With iteration-level batching, LLM serving engines share GPU compute across requests~\citep{vllm}. After prefill builds the KV cache, decoding generates tokens sequentially. When eligible requests exceed the available decode slots, priorities determine which requests advance in that iteration. Output length provides a proxy for decoding work, although batch composition and context length also matter. Numerical length estimates remain useful: countdown priorities subtract generated tokens from the predicted total, and KV cache planning uses estimates of future sequence length.

For request selection, length estimates can support effective priorities even when individual predictions contain errors. Distinguishing a 50-token request from a 500-token request may matter more than refining an estimate from 53 to 50 tokens. The value of further accuracy gains depends on how they change scheduling decisions and what they cost to compute. This motivates a lightweight prediction head that supports useful countdown priorities without excessive prediction overhead. We assess its prediction quality together with the serving benefits and costs it produces.

\begin{takeaway}[title={Takeaway 2: Predict for scheduling}]
Numerical length estimates support remaining-work priorities, but lower prediction error alone does not establish better serving performance. Further accuracy gains should be evaluated against their scheduling benefits and execution costs.
\end{takeaway}

\subsection{Q3: Why encode the prompt twice?}

Many length predictors run a separate text encoder over the prompt~\citep{ssjf,ltr,tie}. For decode scheduling, however, the serving model has already processed that prompt during prefill. Its hidden state reflects both the input text \emph{and} the model that will generate the answer. This raises the possibility of reading the length signal from the representation already used for generation.

Representation studies support this possibility: low-capacity probes recover entity properties and relations from contextual states~\citep{implicitmeaning}, while circuit tracing and Jacobian-lens analyses expose intermediate computations and plans for future output~\citep{biology,globalworkspace}.

Prior work also demonstrates length prediction from hidden states~\citep{trail,star,egtp,estp}. Our design prioritizes obtaining useful scheduling scores with minimal feature processing and predictor computation.

\begin{takeaway}[title={Takeaway 3: Reuse the representation}]
Prefill representations already contain useful length signals. A separate prompt encoder is unnecessary for post-prefill scheduling.
\end{takeaway}

\subsection{Q4: Why spend GPU resources to save GPU resources?}

Length-aware scheduling aims to make better use of limited GPU resources. Yet a GPU-based length predictor consumes memory and compute to decide how those same resources should be allocated. Its cost can offset the benefit of the scheduling decisions it enables.

While GPU decoding is busy, host CPUs may have spare capacity between tokenization, routing, and bookkeeping tasks. Moving prediction to the CPU can use this capacity, but a slow predictor leaves the scheduler waiting or acting on stale information. The relevant cost therefore includes feature transfer and score publication as well as predictor computation.

\begin{takeaway}[title={Takeaway 4: Make CPU prediction practical}]
CPU prediction and initial batch ordering can finish before GPU prefill completes (Section~\ref{par:prefill-overlap}), keeping predictor computation off the GPU.
\end{takeaway}

\section{Method}
\label{sec:method}

\subsection{Overview}
\label{sec:overview}

\method{LEAP} has three core components:

\begin{enumerate}[topsep=0pt,partopsep=0pt,itemsep=6pt,parsep=0pt]
\item \textbf{Linear readout} (Section~\ref{sec:readout}): A lightweight head that extracts length signals from a single prefill hidden state using square-root target regression.
\item \textbf{Online adaptation} (Section~\ref{sec:adapt}): Test-time updates via recursive least squares when domain or task shifts degrade prediction quality.
\item \textbf{Asynchronous CPU execution} (Section~\ref{sec:system}): Prediction and initial batch ordering overlap the remaining GPU prefill, keeping computation off the GPU.
\end{enumerate}

Figure~\ref{fig:architecture} illustrates the request flow. When a substantial domain or task shift weakens the frozen readout, \method{LEAP} adapts by updating the linear coefficients from completion feedback, without retraining the serving model or modifying feature standardization.

\subsection{Mapped-length linear readout}
\label{sec:readout}

We reuse hidden states (Q3) to predict lengths for request ordering (Q2). Let $h_i$ be the final prompt token's hidden state at a selected late, nonfinal layer, and $x_i$ its standardized feature. The coordinatewise mean and scale are fitted on source-training features and remain frozen during serving.

\begin{figure}[!htbp]
\centering
\includegraphics[width=\linewidth]{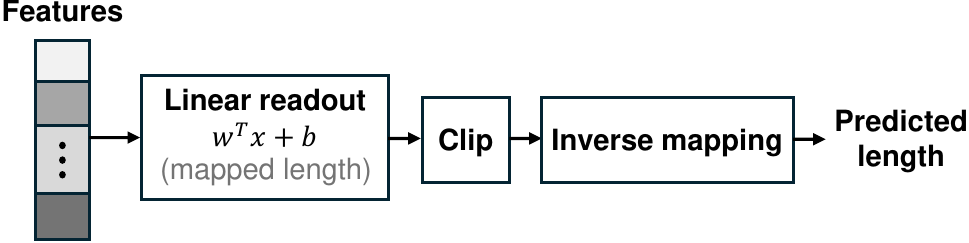}
\par\vspace{-3pt}
\caption{Frozen linear readout with square-root length mapping.}
\label{fig:readout-component}
\vspace{-2pt}
\end{figure}

We use Ridge regression to predict a transformed length target $g(L)$ from $x$. Given $n$ source training pairs $(x_i,y_i)$, where $y_i$ is the target length, the base predictor solves
\begin{equation}
  (w_0,b_0)=\argmin_{w,b}
  \sum_{i=1}^{n}(w^\top x_i+b-g(y_i))^2
  +\lambda_0\|w\|_2^2.
  \label{eq:ridge}
\end{equation}
Here $\lambda_0>0$ regularizes the weights while the intercept remains unpenalized.

\begin{samepage}
The linear readout $z_0(x)=w_0^\top x+b_0$ produces a value on the transformed scale. For an output budget $M$, we clip this value to the valid length range and apply the inverse mapping:
\begin{equation}
  \widehat L_0(x)=g^{-1}\!\left(
    \clip\bigl(z_0(x),g(1),g(M)\bigr)\right).
  \label{eq:inverse}
\end{equation}
\end{samepage}
At deployment, the readout uses folded coefficients $\bar w_0,\bar b_0$ to operate directly on $h$, requiring one dot product followed by clipping and the inverse mapping.

\paragraph{Why square-root mapping?}
\phantomsection
\label{sec:mapping}

The target mapping controls which length errors receive more weight during training. On the original scale, squared loss gives extreme lengths disproportionate influence. A logarithmic mapping compresses the tail, but it also penalizes a fixed token error much less on a long request than on a short one. A square-root mapping weakens this dependence on length. For an error $e$ small relative to length $L$:
\begin{align}
  \bigl(\log(L+e)-\log L\bigr)^2
    &\simeq \frac{e^2}{L^2}, \label{eq:logloss}\\
  \bigl(\sqrt{L+e}-\sqrt L\bigr)^2
    &\simeq \frac{e^2}{4L}. \label{eq:sqrtloss}
\end{align}
Thus, the weight on a fixed absolute error decreases as $1/L^2$ under the logarithm and as $1/L$ under the square root. The square root retains more sensitivity to differences between long requests while still reducing tail dominance.

We use $g(L)=\sqrt L$ as the default compromise between tail compression and long-request separation. The readout is trained on square-root targets, yielding
\begin{equation}
  \widehat L_0(x)=
    \left[\clip\bigl(w_0^\top x+b_0,1,\sqrt M\bigr)\right]^2.
  \label{eq:sqrtpred}
\end{equation}
Clipping before squaring prevents large negative coordinates from becoming large positive length estimates. Figure~\ref{fig:readout-component} illustrates the readout and length conversion for square-root targets.
The choice of map reflects the prediction task and constraints. Log targets prioritize short-output fidelity, while square root compresses long-output differences less aggressively and admits a simple inverse. Alternatives such as piecewise or power-law maps could be explored; with cached training features, switching maps requires only refitting the small head.

\subsection{Online adaptation from completion feedback}
\label{sec:adapt}

\begin{figure}[!htbp]
\centering
\includegraphics[width=\linewidth]{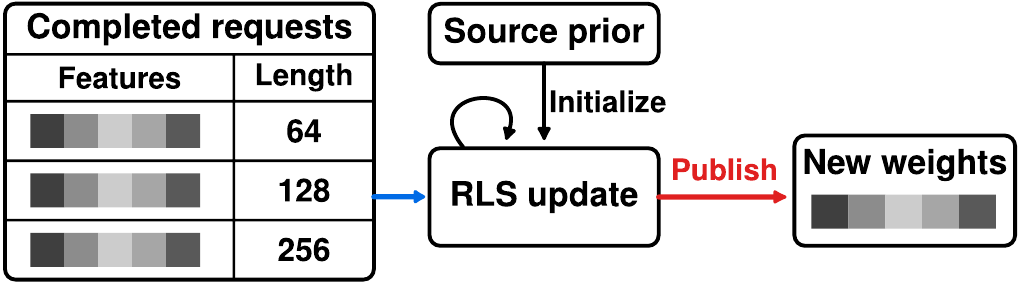}
\par\vspace{-3pt}
\caption{Source-initialized RLS and asynchronous weight publication. Lengths are illustrative.}
\label{fig:adaptation-component}
\vspace{-2pt}
\end{figure}

\paragraph{Incorporating completion feedback.}
To address domain and task shifts (Q1), we update the readout from completion feedback. Hidden states provide model-conditioned length signals, while stochastic decoding can produce different lengths for the same prompt~\citep{egtp}. Source-generation labels train the readout on these length statistics; actual completion labels give RLS a mechanism for recalibrating the relationship under the deployed generation policy.

Each completed request supplies its cached prefill feature $u=[x;1]\in\R^{d+1}$ and mapped-length label $g(L)=\sqrt L$. We fit completion feedback while anchoring the coefficients to the deployed source head, keeping source standardization and the target mapping fixed. Recursive least squares (RLS) incorporates new labels incrementally. Starting from the source head $\theta_0=[w_0;b_0]$, we initialize the inverse source-feature precision:
\begin{equation}
  A_0=U_s^\top U_s+\lambda_{\mathrm p} D,\qquad P_0=A_0^{-1},
  \label{eq:rls-prior}
\end{equation}
where $U_s$ contains source-training features, $\lambda_{\mathrm p}>0$ sets the prior precision, and $D=\operatorname{diag}(1,\ldots,1,0)$ leaves the intercept unpenalized. For the incorporated completions $\mathcal F_t$, the objective is
\begin{equation}
\begin{split}
  \theta_t=\argmin_\theta\;&(\theta-\theta_0)^\top A_0(\theta-\theta_0)\\
  &+\sum_{j\in\mathcal F_t}(u_j^\top\theta-g(L_j))^2.
\end{split}
  \label{eq:adapt}
\end{equation}
The source-feature precision controls how strongly feedback changes the deployed coefficients. RLS minimizes this objective incrementally; when the source head also satisfies the normal equations defined by $A_0$, it is equivalent to joint source-plus-feedback Ridge refitting.

\paragraph{Incremental updates.}
Once features and observed lengths have been matched for $q$ completed requests, they form a batch with feature matrix $U$ and mapped targets $y$. The Woodbury identity gives
\begin{align}
  V&=PU^\top,\qquad G=V(I_q+UV)^{-1}, \label{eq:rls-gain}\\
  \theta^+&=\theta+G(y-U\theta), \label{eq:rls-update}\\
  P^+&=P-GV^\top. \label{eq:rls-covariance}
\end{align}
The gain $G$ maps the new prediction residual to a coefficient correction, while $(\theta,P)$ retains the accumulated information without replaying completed features or computing gradients. Each batch solves a $q\times q$ system and costs $O(qd^2+q^2d+q^3)$ arithmetic. The inverse precision requires $O(d^2)$ CPU state; caching features for $N$ active requests adds $O(Nd)$ space.

\paragraph{Why the recursive update is exact.}
Let $U_f$ stack the incorporated feedback features and let $y_f$ contain their mapped lengths. Expanding Equation~\ref{eq:adapt} gives the normal equations
\begin{equation}
\begin{split}
 A_t&=A_0+U_f^\top U_f,\\
 c_t&=A_0\theta_0+U_f^\top y_f,\qquad \theta_t=A_t^{-1}c_t.
\end{split}
\label{eq:anchored-normal-main}
\end{equation}
The regularized source precision is positive definite, so the minimizer is unique. Suppose $P=A^{-1}$ and $\theta=Pc$ before a new batch. Adding that batch gives $A^+=A+U^\top U$ and $c^+=c+U^\top y$. For $S=I_q+UPU^\top$, Woodbury yields
\begin{equation}
 P^+=P-PU^\top S^{-1}UP,\qquad P^+U^\top=G.
\label{eq:anchored-woodbury-main}
\end{equation}
Consequently,
\begin{equation}
\begin{split}
 \theta^+&=P^+(c+U^\top y)\\
         &=(P-GUP)c+Gy=\theta+G(y-U\theta).
\end{split}
\label{eq:anchored-recursion-main}
\end{equation}
The initialization satisfies these identities, and each batch preserves them. Thus Equations~\ref{eq:rls-gain}--\ref{eq:rls-covariance} solve the anchored objective for every incorporated feedback set in exact arithmetic. Batch size controls update timing; the solution for a fixed feedback set is unchanged. Joint refitting on the mapped source labels $y_s$ additionally requires $A_0\theta_0=U_s^\top y_s$; the anchored guarantee does not require that consistency condition.

\paragraph{CPU execution.}
A separate CPU thread updates $(\theta,P)$ and hands coefficient snapshots to the prediction worker, which rescores cached live features:
\begin{equation}
  z_t(x)=u^\top\theta_t.
  \label{eq:updated}
\end{equation}
Equation~\ref{eq:inverse} converts each score back to a token-length estimate. The source head supplies initial predictions; published RLS versions revise them as feedback arrives. This design keeps matrix updates off the initial-readout queue and reuses prefill features without another GPU forward pass. The scheduler continues using published estimates while an update is pending (Section~\ref{sec:system}).
The prototype incorporates eight completions per update. Only completed requests supply labels, and pending batches do not change the published predictor. At $d=4096$, the inverse precision occupies about 64\,MiB in FP32; the small update system is solved in FP64. Asynchrony removes an update dependency from prediction but does not eliminate CPU or memory-bandwidth costs.

\begin{table*}[!t]
\caption{Initial length prediction across models and datasets. MAE is in tokens; $\rho$ is Spearman correlation. Bold marks the best value.}
\label{tab:prediction-quality}
\centering
\small
\setlength{\tabcolsep}{2pt}
\renewcommand{\arraystretch}{1.08}
\begin{tabularx}{\textwidth}{@{}c*{11}{C}@{}}
\toprule
& \multicolumn{11}{c}{Method} \\
\cmidrule(lr){2-12}
Dataset & \multicolumn{2}{c}{\textbf{\method{LEAP}} (ours)} & \multicolumn{2}{c}{\method{TIE}} & \multicolumn{2}{c}{\method{TRAIL}} & \method{LTR} & \multicolumn{2}{c}{\method{SSJF}} & \multicolumn{2}{c}{\method{EGTP}} \\
\cmidrule(lr){2-3}\cmidrule(lr){4-5}\cmidrule(lr){6-7}\cmidrule(lr){8-8}\cmidrule(lr){9-10}\cmidrule(lr){11-12}
 & MAE $\downarrow$ & $\rho\uparrow$ & MAE $\downarrow$ & $\rho\uparrow$ & MAE $\downarrow$ & $\rho\uparrow$ & $\rho\uparrow$ & MAE $\downarrow$ & $\rho\uparrow$ & MAE $\downarrow$ & $\rho\uparrow$ \\
\midrule
\multicolumn{12}{c}{Qwen3-8B} \\
Alpaca & \textbf{71.5} & \textbf{0.955} & 161.8 & 0.875 & 114.0 & 0.887 & 0.863 & 139.6 & 0.847 & 241.9 & 0.482 \\
ShareGPT & \textbf{171.6} & \textbf{0.900} & 316.1 & 0.675 & 278.8 & 0.719 & 0.733 & 280.4 & 0.726 & 428.7 & 0.142 \\
LMSYS & \textbf{127.5} & \textbf{0.932} & 193.7 & 0.848 & 213.1 & 0.791 & 0.817 & 210.2 & 0.811 & 395.5 & 0.448 \\
\midrule
\multicolumn{12}{c}{Llama-3.1-8B-Instruct} \\
Alpaca & \textbf{63.0} & \textbf{0.942} & 130.7 & 0.821 & 67.9 & 0.933 & 0.862 & 98.1 & 0.865 & 202.2 & 0.462 \\
ShareGPT & \textbf{145.7} & \textbf{0.850} & 214.3 & 0.672 & 160.2 & 0.808 & 0.691 & 196.0 & 0.699 & 276.2 & 0.101 \\
LMSYS & \textbf{105.0} & \textbf{0.922} & 177.3 & 0.768 & 135.1 & 0.841 & 0.782 & 149.6 & 0.785 & 287.8 & 0.268 \\
\midrule
\multicolumn{12}{c}{Mistral-7B-Instruct-v0.3} \\
Alpaca & \textbf{65.4} & \textbf{0.931} & 105.5 & 0.833 & 81.5 & 0.884 & 0.833 & 90.5 & 0.852 & 145.2 & 0.550 \\
ShareGPT & \textbf{143.9} & \textbf{0.784} & 193.4 & 0.618 & 173.9 & 0.667 & 0.640 & 178.3 & 0.646 & 243.1 & 0.243 \\
LMSYS & \textbf{106.9} & \textbf{0.896} & 163.5 & 0.792 & 139.7 & 0.808 & 0.805 & 131.9 & 0.808 & 198.9 & 0.500 \\
\bottomrule
\end{tabularx}
\end{table*}

\subsection{Asynchronous prediction and scheduling}
\label{sec:system}

To address the resource tradeoff in Q4, we execute prediction asynchronously on the CPU alongside GPU inference.

\begingroup
\setlength{\intextsep}{2pt}
\begin{figure}[!htbp]
\centering
\includegraphics[width=\linewidth]{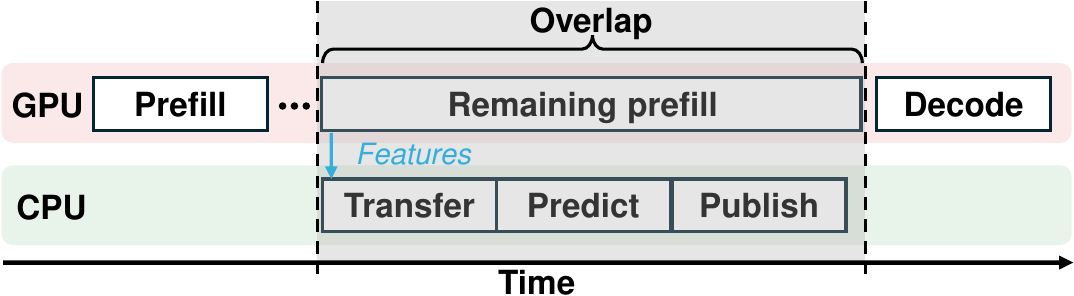}
\par\vspace{-5pt}
\caption{CPU feature transfer, prediction, and publication overlap the remaining GPU prefill. Timing is schematic.}
\label{fig:scheduling-component}
\vspace{-7pt}
\end{figure}
\endgroup

At the selected layer, we capture the final prompt token's hidden state and transfer it to the CPU while the GPU continues prefill. Once the transfer completes, a CPU worker applies the current predictor and publishes the length estimate. After the scheduler consumes the published estimate, subsequent eligible decode decisions use the revised order. The CPU worker can finish prediction and initial batch ordering before GPU prefill completes, as verified by the GPU readiness probe in the \hyperref[par:prefill-overlap]{prefill-overlap experiment}. The engine applies the prepared order at its next eligible decode decision. The feature is captured and transferred only once per request.

At decode step $t$, request $i$ has generated $a_i(t)$ tokens and has predicted total length $\widehat L_i(t)$. Its remaining-length score is
\begin{equation}
  s_i(t)=\max\{1,\widehat L_i(t)-a_i(t)\}.
  \label{eq:remaining}
\end{equation}

When eligible KV-resident requests exceed decode capacity, \method{LEAP} selects the advancing subset by this score; deferred requests retain their KV state. The engine controls admission, prefill, memory allocation, and fairness overrides.

All methods within each comparison share waiting protection: a 128-step guard in the live engine and \method{TIE}'s exponential Aging rule for cached replays and the LMSYS load control~\citep{tie} (Appendix~\ref{app:matched-aging}).

Countdown updates the remaining-length score at each decoding step; RLS revises the predicted total length after a batch of completions. The scheduler consumes newly published estimates when available, while countdown continues without waiting for the update thread. Until the first prediction is ready, requests use their output budget as the total-length estimate. If updates lag, existing estimates continue counting down.

Published scores carry request identity and predictor-version metadata.
The scheduler accepts only results for matching live requests, preventing stale results from changing unrelated priorities.
Score-delivery measurements therefore follow an estimate through publication to scheduler consumption, capturing the delay beyond the readout itself.

Completion feedback is matched by request identity and incorporated in full batches of eight. A final incomplete batch is not fitted, and each incorporated label is the observed served length under the request's output budget.

\FloatBarrier

\section{Evaluation}
\setlength{\intextsep}{10pt}
\setlength{\parskip}{4pt}
\label{sec:evaluation}

\subsection{Experimental Setup}

\paragraph{Datasets and models.}
We evaluate Qwen3-8B, Llama-3.1-8B-Instruct, and Mistral-7B-Instruct-v0.3 on instruction-following requests from Alpaca~\citep{alpaca} and conversations from ShareGPT~\citep{sharegpt} and LMSYS-Chat-1M~\citep{lmsysdata}. Frozen LMSYS predictors transfer to Alpaca, MATH~\citep{mathdata}, and non-English Aya~\citep{ayadata} requests to test task and language shifts. Methods share request splits, with prompt and conversation-parent overlap excluded between training and test. The first generated output defines each test label.

\paragraph{Baselines.}
We compare against uncertainty-aware \method{TIE}~\citep{tie}, rank-based \method{LTR}~\citep{ltr}, proxy predictor \method{SSJF}~\citep{ssjf}, and hidden-state predictors \method{TRAIL}~\citep{trail} and \method{EGTP}~\citep{egtp}. \method{FCFS} supplies the no-prediction scheduling control. Initial-quality comparisons use \method{TIE}'s bounded expectation and \method{TRAIL}'s prefill head; serving comparisons include \method{TRAIL}'s arrival predictor, initial and online heads, and probability filter. \method{SSJF} uses source-training target normalization, inverted at inference.

\paragraph{Training.}
The default Qwen readout uses the final prompt token at block 30, square-root targets, and frozen source-training standardization. \method{LEAP} fits its head on 2,880 training requests. Baseline fitting retains each author's objective and checkpoint-selection rule. \method{LEAP} and \method{TIE} use repeated-generation targets, whereas \method{LTR}, \method{SSJF}, and \method{EGTP} use first-rollout labels. Readout regularization minimizes token MAE on source validation data, without target-domain tuning. Prediction comparisons use one training/generation seed unless noted.

\paragraph{Serving and metrics.}
Serving runs on one A800 80GB PCIe GPU and eight Xeon Gold 6348 logical CPUs using vLLM 0.8.5.post1 in eager mode. \method{LEAP} executes on CPU; deep predictors share the GPU. Full-cost comparisons replay common requests and recorded token trajectories under one scheduling policy, so only the predictor path differs. The engine uses FIFO prefill, 512 resident requests, 64 decode slots, and a 128-step waiting guard; requests without an accepted score fall back to \method{FCFS}. Test labels are first-rollout lengths capped at 2,048 tokens.
We report MAE and Spearman $\rho$ for prediction (\method{LTR}'s ordinal scores have no token MAE), and mean/P90 per-token latency (PTL), time to first token (TTFT), and throughput for serving. Latency starts at planned arrival and includes prediction, publication, queueing, and execution; PTL divides completion latency by actual output tokens.

\subsection{Prediction Quality and Domain Transfer}
\label{sec:prediction-transfer}

\indent\textbf{Prediction from prefill representations.}
Table~\ref{tab:prediction-quality} compares initial length estimates across three models and three datasets.

\indent\textbf{Analysis.}
\method{LEAP} has the lowest MAE and highest Spearman $\rho$ in all nine settings, reducing MAE by 7--39\% relative to the best baseline in each setting, without a separate prompt encoder.

\begin{questionanswer}
\textbf{Answer to Q3:} Reusing a single prefill hidden state with a linear readout avoids a second prompt encoder and yields lower MAE and higher Spearman correlation than the evaluated independent-encoder predictors in all nine settings (Table~\ref{tab:prediction-quality}).
\end{questionanswer}

\indent\textbf{Static versus dynamic prediction.}
\phantomsection
\label{sec:dynamic-control}
Refreshing estimates during decoding can lower error, but at a runtime cost. Table~\ref{tab:dynamic-sufficiency} compares static countdown (-S) with hidden-state refreshes during decoding (-D). MAE is measured at matched live token positions on 1,000 test requests per domain; $\Delta$Resp. comes from paired serving runs on 128 held-out requests per domain that share initial estimates and token trajectories, with the full prediction pipeline included.

\begin{table}[!htbp]
\caption{Prediction accuracy and serving cost. Green rows use static countdown (-S); orange rows refresh hidden states during decoding (-D). $\Delta$Resp. is relative to the static version.}
\label{tab:dynamic-sufficiency}
\centering
\small
\setlength{\tabcolsep}{2pt}
\renewcommand{\arraystretch}{1.10}
\begin{tabularx}{\linewidth}{@{}CCCCC@{}}
\toprule
& \multicolumn{2}{c}{ShareGPT} & \multicolumn{2}{c}{LMSYS} \\
\cmidrule(lr){2-3}\cmidrule(lr){4-5}
Predictor & MAE & \mbox{$\Delta$Resp. (\%)} & MAE & \mbox{$\Delta$Resp. (\%)} \\
\midrule
\rowcolor{LIEStaticRow}
\method{TRAIL-S} & 244.67 & --- & 170.69 & --- \\
\rowcolor{LIEDynamicRow}
\method{TRAIL-D} & 167.01 & +12.89 & 108.84 & +9.16 \\
\rowcolor{LIEStaticRow}
\method{LEAP-S} & 154.60 & --- & 111.92 & --- \\
\rowcolor{LIEDynamicRow}
\method{LEAP-D} & 141.96 & +9.42 & 103.82 & +8.37 \\
\bottomrule
\end{tabularx}
\end{table}

For \method{LEAP}, static and dynamic heads use repeated-generation and round-0 remaining-length targets, respectively, so this comparison does not isolate feature refresh alone. RLS parameter adaptation is separate from these decode-time refreshes.

\begin{questionanswer}
\textbf{Answer to Q2:} Lower prediction error does not necessarily translate into lower serving latency. In our paired controls, dynamic prediction improves MAE but increases mean response time, supporting a lightweight static predictor that preserves useful ordering at low execution cost.
\end{questionanswer}

\begin{figure}[!htbp]
\centering
\includegraphics[width=\linewidth]{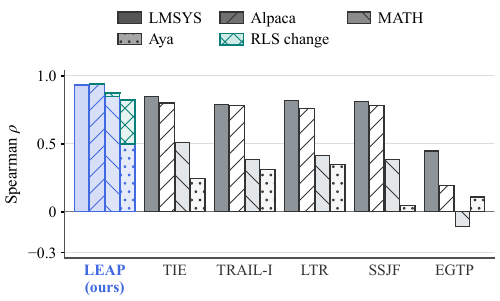}
\caption{Ranking after LMSYS transfer on Qwen3-8B. Crosshatched extensions show RLS changes to \method{LEAP}.}
\label{fig:ood-bars}
\end{figure}

\indent\textbf{Testing frozen cross-domain generalization.}
We transfer frozen LMSYS predictors to Alpaca, MATH, and Aya, and measure how much RLS feedback recovers (Figure~\ref{fig:ood-bars}). Alpaca feedback requests are disjoint from its test set; on MATH and Aya, each request is predicted before its label enters RLS.

\begin{samepage}
\indent\textbf{Analysis.}
The frozen readout transfers well to Alpaca and MATH, but the language shift to Aya degrades its ranking. On Aya, RLS raises Spearman $\rho$ from 0.498 to 0.820, a 64.7\% relative improvement, by updating only the readout, without touching the backbone.
\end{samepage}

\begin{questionanswer}
\textbf{Answer to Q1:} Generalization is useful but not uniform: frozen \method{LEAP} retains useful rankings on Alpaca and MATH, while completion feedback substantially restores ranking quality on Aya (Figure~\ref{fig:ood-bars}).
\end{questionanswer}

\subsection{Serving on Similar Workloads}
\label{sec:similar-serving}
\label{sec:representative-serving}

\indent\textbf{Comparing predictors with full deployment costs.}
Table~\ref{tab:single30-serving} reports ShareGPT serving with all prediction costs included. Requests arrive at 2.65 requests/s and replay the same output trajectories under every predictor.

\begin{table}[!htbp]
\caption{Serving performance on ShareGPT under a shared selector. PTL: ms/token; TTFT: ms. Rows are ordered from higher to lower mean PTL. Our results are bold.}
\label{tab:single30-serving}
\centering
\scriptsize
\setlength{\tabcolsep}{2pt}
\renewcommand{\arraystretch}{1.12}
\begin{tabularx}{\linewidth}{lCCCCC}
\toprule
Method & \shortstack{Mean\\PTL} & \shortstack{P90\\PTL} & \shortstack{Mean\\TTFT} & \shortstack{P90\\TTFT} & Tok/s \\
\midrule
\method{FCFS} & 250.97 & 359.96 & 55.10 & 74.85 & 1094.03 \\
\method{TRAIL} & 107.42 & 189.32 & 67.18 & 90.28 & 944.71 \\
\method{TIE} & 78.48 & 152.33 & 66.50 & 85.22 & 1089.84 \\
\method{SSJF} & 76.07 & 154.50 & 62.27 & 80.34 & 1108.46 \\
\method{LTR} & 75.62 & 151.95 & 60.40 & 80.17 & 1120.88 \\
\midrule
\method{LEAP} (ours) & \textbf{63.58} & \textbf{127.47} & \textbf{55.27} & \textbf{73.43} & \textbf{1141.93} \\
\bottomrule
\end{tabularx}
\end{table}

\begin{samepage}
\indent\textbf{Analysis.}
\method{LEAP} lowers mean PTL by 15.9--40.8\% across the evaluated predictors: 15.9\% relative to \method{LTR}, 19.0\% to \method{TIE}, and 40.8\% to \method{TRAIL}. \method{LTR} is the strongest baseline here by both mean PTL and throughput; \method{LEAP} also raises throughput by 1.9\% over \method{LTR}, while keeping TTFT close to \method{FCFS}. Long requests still wait longer than under \method{FCFS}, the usual cost of favoring short work. Results are single runs on a shared host.
\end{samepage}

\begin{figure}[!htbp]
\centering
\includegraphics[width=\linewidth]{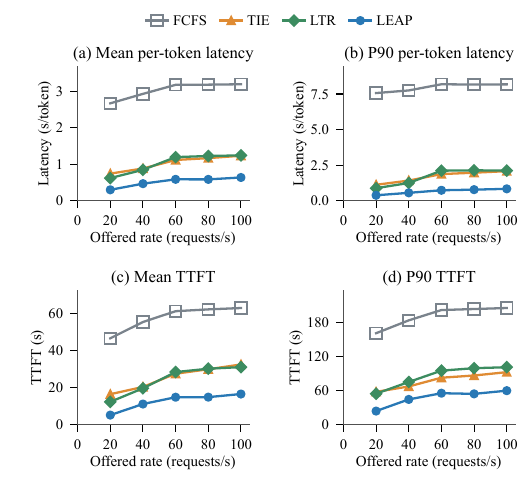}
\caption{LMSYS serving latency versus offered rate under shared post-prefill scheduling. All four methods serve the same 1,024 requests and token trajectories.}
\label{fig:lmsys-rps4200}
\end{figure}

\indent\textbf{Prediction cost under rising load.}
We replay the same 1,024 held-out LMSYS requests at 20--100 requests/s (Figure~\ref{fig:lmsys-rps4200}). All methods share request identities, token trajectories, and one post-prefill scheduling rule, so they differ only in the length estimate and where it is computed. \method{LEAP} predicts on CPU, while \method{TIE} and \method{LTR} run on the serving GPU; client latency includes this prediction cost. Protocol details are in Appendix~\ref{app:lmsys-rps4200}.

\indent\textbf{Analysis.}
\method{LEAP} has the lowest value in all four panels at every rate, and its mean per-token latency stays at about half that of the better GPU predictor as load rises. All three predictors stay far below \method{FCFS}, so length-aware ordering accounts for most of the gain over \method{FCFS}. The remaining gap combines \method{LEAP}'s more accurate ranking (Table~\ref{tab:prediction-quality}) with keeping prediction off the serving GPU; this single-run experiment does not separate the two.

\subsection{Scheduling under Larger Domain Shifts}
\label{sec:shift-serving}

\indent\textbf{Scheduling after domain transfer.}
We replay Aya and MATH streams in simulation with cached predictions from frozen LMSYS models. All methods share the remaining-length priority, the Aging rule in Equation~\ref{eq:aging}, and the same admission and execution models. Table~\ref{tab:matched-aging} reports mean response time at relative offered loads 1.2 and 1.5. RLS sees each label only after the simulated completion. Predictor runtime costs are excluded here; Section~\ref{sec:similar-serving} covers them.

\begin{table}[H]
\caption{Simulated mean response time after domain transfer with matched Aging (s; lower is better). Bold: ours.}
\label{tab:matched-aging}
\centering
\small
\setlength{\tabcolsep}{3pt}
\renewcommand{\arraystretch}{1.08}
\begin{tabularx}{\linewidth}{@{}c*{4}{C}@{}}
\toprule
& \multicolumn{2}{c}{Aya} & \multicolumn{2}{c}{MATH} \\
\cmidrule(lr){2-3}\cmidrule(lr){4-5}
Method & $1.2$ & $1.5$ & $1.2$ & $1.5$ \\
\midrule
\method{SSJF} & 154.83 & 253.29 & 80.30 & 126.83 \\
\method{EGTP} & 136.62 & 223.09 & 96.54 & 157.83 \\
\method{FCFS} & 138.40 & 226.71 & 91.66 & 148.90 \\
\method{TRAIL} & 125.02 & 201.67 & 77.52 & 123.67 \\
\method{LTR} & 123.32 & 197.28 & 75.12 & 118.93 \\
\method{TIE} & 119.56 & 190.06 & 75.05 & 118.34 \\
\midrule
\method{LEAP} (ours) & \textbf{115.18} & \textbf{180.59} & \textbf{67.39} & \textbf{100.94} \\
\method{LEAP}+RLS (ours) & \textbf{108.07} & \textbf{161.18} & \textbf{64.83} & \textbf{94.36} \\
\bottomrule
\end{tabularx}
\end{table}

\indent\textbf{Analysis.}
With protection held fixed, latency differences come from ranking alone, and they track the ranking gaps in Figure~\ref{fig:ood-bars}. Frozen \method{LEAP}'s advantage over \method{TIE} is largest on MATH (up to 14.7\%), and all gaps widen at the higher load, where longer queues make ordering matter more. RLS helps most on Aya, where the frozen ranking is weakest. Better ranking does not guarantee better tails: RLS raises P95 on MATH, and \method{LEAP}, \method{LEAP}+RLS, and \method{TIE} all keep a higher P95 than \method{FCFS}.
\par

\subsection{Backbone Scale}

\indent\textbf{Testing backbone scale.}
We fit the same square-root readout on dense Qwen3 models from 0.6B to 32B, using single-rollout labels and one fixed layer per model (Table~\ref{tab:qwen-scale}). Each checkpoint is evaluated on its own outputs.

\begin{table}[!htbp]
\caption{Length prediction across Qwen3 model sizes.}
\label{tab:qwen-scale}
\centering
\small
\setlength{\tabcolsep}{2pt}
\renewcommand{\arraystretch}{1.12}
\begin{tabularx}{\linewidth}{CCCCC}
\toprule
Size & \multicolumn{2}{c}{ShareGPT} & \multicolumn{2}{c}{LMSYS} \\
\cmidrule(lr){2-3}\cmidrule(lr){4-5}
 & MAE & $\rho$ & MAE & $\rho$ \\
\midrule
0.6B & 151.38 & 0.836 & 94.81 & 0.910 \\
1.7B & 206.26 & 0.856 & 144.07 & 0.921 \\
4B & 184.84 & 0.879 & 124.81 & 0.932 \\
8B & 173.26 & 0.901 & 129.14 & 0.936 \\
14B & 160.31 & 0.909 & 116.13 & 0.936 \\
32B & 155.30 & 0.915 & 116.68 & 0.938 \\
\bottomrule
\end{tabularx}
\end{table}

\indent\textbf{Analysis.}
Spearman $\rho$ stays above 0.83 at every size, so the readout ports across checkpoints. Long-output ordering is unreliable in the 0.6B model, so a high aggregate $\rho$ does not ensure uniform quality across length groups. Because outputs differ per checkpoint, this does not isolate the effect of size.

\subsection{Target-map Ablation}
\label{sec:map-ablation}

\indent\textbf{Testing short- and long-output accuracy.}
Table~\ref{tab:single30-maps} compares target maps on the same features, averaged over ShareGPT and LMSYS. Short requests have at most 256 output tokens and long requests exceed 1,024; ``All'' also includes medium requests.

\begin{table}[!htbp]
\caption{Effect of the target map on length prediction (mean MAE / Spearman $\rho$).}
\label{tab:single30-maps}
\centering
\footnotesize
\setlength{\tabcolsep}{3pt}
\begin{tabularx}{\linewidth}{@{}cCCC@{}}
\toprule
Map & Short & Long & All \\
\midrule
$\log L$ & 50.7 / 0.7907 & 310.4 / 0.5832 & 155.9 / 0.9122 \\
$\sqrt L$ & 68.5 / 0.7179 & 279.3 / 0.6009 & 149.6 / 0.9161 \\
$L$ & 105.4 / 0.4881 & 266.3 / 0.6097 & 162.7 / 0.8984 \\
\bottomrule
\end{tabularx}
\end{table}

\indent\textbf{Analysis.}
Log gives lower prediction error and better ordering on short requests. Square root compresses long lengths less aggressively, improving error and ordering for long requests while avoiding the larger short-request errors of identity. We therefore use square root by default.

\subsection{Single-layer versus All-layer Readout}
\label{sec:input-width}

Collecting all layers may improve prediction but adds collection, buffering, and transfer. Table~\ref{tab:input-width} compares block 30 with the concatenation of all hidden states on Qwen3-8B, using the same target, Ridge fit, and splits, averaged over ShareGPT and LMSYS.

\begin{table}[!htbp]
\caption{Single-layer versus all-layer length prediction on Qwen3-8B.}
\label{tab:input-width}
\centering
\footnotesize
\setlength{\tabcolsep}{4pt}
\begin{tabularx}{\linewidth}{@{}lCC@{}}
\toprule
Input & Mean MAE & Mean Spearman $\rho$ \\
\midrule
Block 30 & 149.53 & 0.9160 \\
All states & 144.63 & 0.9204 \\
\bottomrule
\end{tabularx}
\end{table}

\indent\textbf{Analysis.}
All layers lower MAE by only 3.3\% and give little mean-response gain in an idealized replay that ignores transfer cost (Appendix~\ref{app:input-width}). One layer shrinks the FP32 input from 592 to 16\,KiB.

\subsection{Prediction and Deployment Costs}
\label{sec:prediction-cost}

\indent\textbf{Measuring predictor storage cost.}
Table~\ref{tab:predictor-cost} reports logical FP32 predictor weights in decimal KB/MB/GB, excluding the serving LLM, activations, cached features, and solver state.

\begin{table}[!htbp]
\caption{Predictor parameter counts and FP32 weight storage.}
\label{tab:predictor-cost}
\centering
\small
\setlength{\tabcolsep}{5pt}
\renewcommand{\arraystretch}{1.12}
\begin{tabularx}{\linewidth}{CCC}
\toprule
Predictor & Parameters & FP32 weights \\
\midrule
\method{TIE} & 185,211,394 & 0.741~GB \\
\method{LTR} & 125,240,064 & 0.501~GB \\
\method{SSJF} & 109,597,313 & 0.438~GB \\
\method{EGTP} & 8,435,732 & 33.74~MB \\
\method{TRAIL} & 2,118,184 & 8.47~MB \\
\method{LEAP} (ours) & \textbf{4,097} & \textbf{16.39~KB} \\
\bottomrule
\end{tabularx}
\end{table}

\begin{samepage}
\indent\textbf{Analysis.}
\method{LEAP} uses $30{,}569\times$ fewer predictor parameters than \method{LTR}, the strongest serving baseline in Table~\ref{tab:single30-serving}, and $45{,}207\times$ fewer than \method{TIE}. Deployment assessment must also account for feature transfer, score publication, and scheduler consumption, which we measure next.
\end{samepage}

\begin{figure}[!htbp]
\centering
\includegraphics[width=\linewidth]{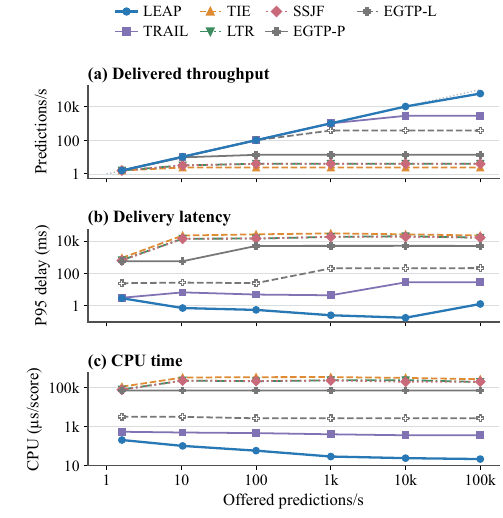}
\caption{CPU score delivery under matched core budgets (batch eight).}
\label{fig:cpu-deployment}
\end{figure}

\indent\textbf{Testing CPU prediction under a fixed budget.}
Figure~\ref{fig:cpu-deployment} measures score delivery with one prediction core per method on matched Qwen3-8B/ShareGPT inputs, excluding serving-model execution. Hidden-state paths include feature transfer and pooling (details: Appendix~\ref{app:current-cpu-cost}).

\indent\textbf{Analysis.}
With one core per method, the CPU cost of each score (panel~(c)) determines both throughput and delay. Text predictors run a full encoder for every request, so a core delivers only a few scores per second and queued scores wait tens of seconds. Hidden-state baselines skip the encoder but still run million-parameter heads, and \method{EGTP}-P must also rebuild logits on CPU, so they saturate between about ten and a few thousand scores per second. \method{LEAP}'s readout is a single 4,097-parameter affine map. One core keeps pace with the offered rate up to 10,000 predictions/s, saturates near 58,000/s, and keeps P95 delay below 3\,ms at every rate.

\begin{figure}[H]
\centering
\includegraphics[width=\linewidth]{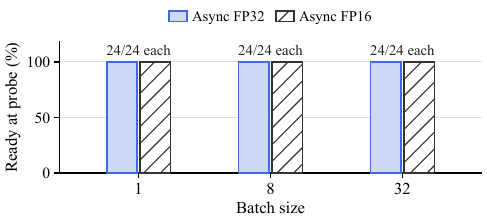}
\caption{Share of trials in which prediction and initial batch order are ready before the final RMSNorm on Qwen3-8B (24 trials per bar).}
\label{fig:prefill-overlap-readiness}
\end{figure}

\indent\textbf{Completing prediction during prefill.}
\phantomsection
\label{par:prefill-overlap}
Figure~\ref{fig:prefill-overlap-readiness} measures readiness during Qwen3-8B prefill. A CPU worker reads block-30 features and sets a flag after prediction and initial batch ordering. A GPU probe checks it before final RMSNorm, while prefill work remains. The post-prefill negative control starts CPU work only after prefill returns (protocol: Appendix~\ref{app:overlap-protocol}).

\indent\textbf{Analysis.}
All 144 asynchronous trials were ready, while all 72 controls were unready. Feature transfer, prediction, and initial ordering finish within the remaining prefill at batch sizes 1, 8, and 32 with FP32/FP16 transport. The probe measures readiness, not the remaining time margin; admission and KV allocation remain with the engine.

\begin{questionanswer}
\textbf{Answer to Q4:} CPU execution sustains higher score-delivery throughput under the tested core budget (Figure~\ref{fig:cpu-deployment}), and prediction and initial ordering finish before prefill ends in all 144 overlap trials (Figure~\ref{fig:prefill-overlap-readiness}).
\end{questionanswer}

\section{Related Work}
\setlength{\parskip}{4pt}
\label{sec:related}

\paragraph{Prediction with auxiliary models.}
\citet{zheng2023length} use response-length estimates to group requests for scheduling. \method{TIE}~\citep{tie} models length uncertainty with a DeBERTa encoder, \method{SSJF}~\citep{ssjf} predicts lengths with a proxy model, and \method{LTR}~\citep{ltr} learns priorities with a separate ranking model. These methods score requests before prefill but require additional model storage and prompt encoding.

\paragraph{Hidden-state prediction and availability.}
\method{TRAIL}~\citep{trail} and \method{STAR}~\citep{star} reuse hidden states but retain MLP prediction heads. \method{TRAIL} already overlaps CPU prediction with later GPU layers, but its CPU path pools all prompt-token states and evaluates an 8.47~MB MLP. These weights exceed a core's private L2 cache, so each batch reloads them from the shared L3 cache or memory used by the serving process, and pooling work grows with prompt length. \method{LEAP}'s 16.39~KB readout stays in a core's private cache, and it consumes one hidden state per request.

\paragraph{Feature aggregation costs.}
\method{EGTP}~\citep{egtp} pools states using token entropy; \method{ESTP}~\citep{estp} adds attention-based semantic weights. Both compute entropy over the full vocabulary, requiring logit projection when logits are unavailable; \method{ESTP} also extracts and aggregates attention weights.

\paragraph{Parameter adaptation.}
Refreshing predictions from new decode states does not itself adapt predictor parameters to a new domain. The MLP heads in \method{TRAIL} and \method{STAR} are trained with gradient-based optimization; refitting them on completion feedback entails further optimization. Self-supervised test-time training~\citep{ttt} likewise updates parameters through a gradient-based auxiliary objective. \method{LEAP} combines a single-state linear readout with gradient-free RLS adaptation and asynchronous CPU execution.

\section{Conclusion}

\method{LEAP} combines cross-domain ranking performance with low-cost prediction and completion-driven updates. The frozen readout generalizes across the evaluated datasets, while RLS restores ranking quality under the larger language shift to Aya. CPU-asynchronous execution turns the compact readout into measured serving benefits. These results connect length prediction to the cost of maintaining useful scheduling decisions as domains and tasks change.

\section{Limitations \& Future Work}
\label{sec:limitations-future}

\textbf{Tail latency and fairness.} Length-aware scheduling prioritizes short requests, which inherently trades mean latency for tail fairness. Introducing waiting-time protections can balance this tradeoff, though methods differ in how strictly they enforce bounds.

\textbf{Decoding policy.} The readout is conditioned only on the prefill representations and does not account for decoding-time choices such as temperature or top-$k$ parameters. Extensions could condition predictions on these factors.

\textbf{Adaptation dynamics.} RLS updates use uniform weights over all completions and can be extended with forgetting factors or selective weighting for special cases such as capped outputs.

\textbf{Serving scope.} The prototype integrates with single-GPU eager-mode vLLM. Multi-GPU, prefix-cached, and chunked-prefill serving are separate concerns.

\bibliographystyle{mlsys2025}
\bibliography{references}

\section*{AI Usage Disclosure}
Generative AI tools, including OpenAI Codex, assisted with method refinement, mathematical exposition, and experimental design. They also assisted with implementation, debugging, experiment execution, data reformatting, and result interpretation. We used generative AI for literature search and summarization, manuscript drafting and language editing, and reference formatting. AI tools also helped create and revise scientific figures and editable graphics. Verification included model and input checks, implementation consistency checks, and comparisons between reported numbers and saved experiment outputs. The authors take responsibility for the final text, methods, figures, code, and reported results.

\clearpage
\appendix
\twocolumn[\section*{Appendix}]
\setlength{\parskip}{0.5pt}
\setlength{\intextsep}{6pt}
\setlength{\textfloatsep}{8pt}
\setlength{\abovedisplayskip}{4pt plus 1pt minus 1pt}
\setlength{\belowdisplayskip}{4pt plus 1pt minus 1pt}
\brokenpenalty=100 
\section{Readout and Adaptation Details}
\label{app:estimation}

\subsection{Base predictor}
\label{app:base-fit}

Source standardization uses coordinatewise training means $\mu$ and scales $s$: $x_i=(h_i-\mu)/s$. These statistics remain frozen during deployment and adaptation. They are absorbed into the deployed base-head coefficients; RLS uses the same standardized features with an appended intercept (Section~\ref{sec:adapt}).

\subsection{Source precision and recursive solver}
\label{app:residual-fit}

The source precision $A_0$ is computed offline from training features only, using the deployed head's frozen mean and scale. With the prior regularization parameterized as $\lambda_{\mathrm p}=\alpha d$, separately from the base-fit $\lambda_0$, the current instantiation uses $\alpha=1$, $d=4096$, and 2,880 source-training requests. The augmented intercept is not regularized. The deployed coefficients are converted to standardized coordinates exactly: if the folded head is $\bar w_0^\top h+\bar b_0$, then $\theta_0=[\bar w_0\odot s;\bar b_0+\mu^\top\bar w_0]$. Thus adaptation starts from the same deployed predictor.

The implemented recursion solves the anchored objective in Equation~\ref{eq:adapt}, as derived in Section~\ref{sec:adapt}. All incorporated completions have equal weight; there is no sliding window or forgetting factor.

The implementation stores $P$ in FP32 and coefficients in FP64. It forms $V$ by an FP32 matrix product, promotes it to FP64 for the small solve and coefficient update, and casts the precision downdate back to FP32. The solve computes $SX=V^\top$, $G=X^\top$, with $S=I_q+UV$; this and Equation~\ref{eq:rls-covariance} rely on symmetry in exact arithmetic. Finite-precision drift must therefore be checked against a direct anchored solve. The implementation checks finite coefficients and the small system; independent numerical comparisons are reported in Appendix~\ref{app:rls-equivalence}.

The current update batch contains $q=8$ completions, with no feature clipping, target discounting, or source-sample replay. The updater consumes full batches in completion order. Shutdown drains queued updates; a final incomplete batch is not fitted. Each label is the observed served length under the output budget, including capped completions; it does not reveal an uncapped latent length.

\subsection{Numerical checks}
\label{app:rls-equivalence}

An FP64 synthetic check compares recursive coefficients with an independently solved joint Ridge system after every full eight-sample batch; the maximum absolute coefficient difference is below $10^{-12}$. Precision diagnostics on two 4,240-request conversation-transfer streams compare FP32 inverse-precision storage and FP64 small solves against an FP64 RLS reference. Without periodic symmetrization, the maximum token-prediction differences are approximately $0.00024$ and $0.00160$. These finite-stream checks measure numerical error, not a universal floating-point bound or serving speedup.

\section{Deployment Configuration and Measurement}
\label{app:deployment}

\subsection{Execution and feedback protocol}

Algorithm~\ref{alg:lifecycle} specifies the request lifecycle. The worker joins completion labels to cached features by request identity and retains early labels until their features arrive. Only matched completions count toward the update batch. Publications carry identity and version metadata, so results for completed or unrelated requests cannot replace live scores. The prototype integrates with vLLM V0 using a native C++ CPU worker.

\begin{algorithm}[!htbp]
\caption{Completion-driven prediction and scheduling}
\label{alg:lifecycle}
\small
\textbf{Notation:} $\mathcal C$: live-feature cache; $\mathcal A_t$: eligible requests; $B$: decode capacity; $\theta_{\rm pub}$: published head. $\ell_\theta(u)$ applies Equations~\ref{eq:updated} and~\ref{eq:inverse}.
\begin{algorithmic}[1]
\REQUIRE $\theta_0,P_0,g,q$
\STATE $(\theta,P)\gets(\theta_0,P_0);\quad \theta_{\rm pub}\gets\theta_0;\quad \mathcal C\gets\varnothing;\quad \mathcal Q\gets[\,]$
\STATE \textbf{on arrival} $(i)$: $\widehat L_i\gets M$
\STATE \textbf{at prefill layer} $(i,h_i)$: \textsc{AsyncD2H}$(i,h_i)$
\STATE \textbf{prediction worker: feature ready} $(i,h_i)$
\STATE \quad $u_i\gets[\textsc{Standardize}(h_i);1];\quad \mathcal C[i]\gets u_i$
\STATE \quad \textsc{PublishInitial}$(i,\ell_{\theta_0}(u_i))$
\STATE \quad \textsc{PublishLatest}$(i,\ell_{\theta_{\rm pub}}(u_i))$
\STATE \textbf{prediction worker: matched completion} $(i,L_i)$
\STATE \quad $\mathcal Q\gets\mathcal Q\mathbin{\|}(\mathcal C[i],g(L_i));\quad \textsc{Erase}(\mathcal C,i)$
\STATE \quad \textbf{if} $|\mathcal Q|=q$ \textbf{then} \textsc{EnqueueUpdate}$(\mathcal Q)$; $\mathcal Q\gets[\,]$
\STATE \textbf{update thread: batch ready} $(U,y)$
\STATE \quad $(\theta,P)\gets\textsc{RLS}(\theta,P,U,y)$ \hfill Eqs.~\ref{eq:rls-gain}--\ref{eq:rls-covariance}
\STATE \quad \textsc{HandoffSnapshot}$(\theta)$
\STATE \textbf{prediction worker: between jobs}
\STATE \quad \textbf{if} new snapshot \textbf{then} $\theta_{\rm pub}\gets\textsc{NewestSnapshot}()$
\STATE \qquad \textsc{PublishLive}$(\mathcal C,\theta_{\rm pub})$ \textbf{end if}
\STATE \textbf{on decode} $(\mathcal A_t,B)$
\STATE \quad $\widehat L\gets\textsc{ConsumeValid}(\widehat L)$
\STATE \quad $s_i\gets\max\{1,\widehat L_i-a_i(t)\},\quad i\in\mathcal A_t$
\STATE \quad \textsc{Advance}$(\textsc{GuardedShortest}(\mathcal A_t,s,B))$
\end{algorithmic}
\end{algorithm}

\subsection{Matched-Aging transfer replay}
\label{app:matched-aging}
Table~\ref{tab:matched-aging} reuses fixed Qwen3-8B output trajectories, taking the first 4,200 eligible requests in the frozen stream order for each of Aya and MATH. The same prefix is used for all methods; requests are not resampled by length. Calibration and the entire queueing process are recomputed for 4,200 requests.
The single arrival seed is 42. Offered load $\rho_{\mathrm{load}}\in\{1.2,1.5\}$ scales the arrival rate relative to the capacity reference obtained from the same-request \method{FCFS} simulation.
All learned predictors share greedy KV admission, the same last-victim rule, a 256-request batch limit, a 380,000-token KV budget, and the same fitted A800 prefill/decode/KV/swap cost model.
The exponential Aging rule ranks requests by
\begin{equation}
  \tilde s_i(t)=s_i(t)\,\gamma^{W_i(t)/\tau},
  \label{eq:aging}
\end{equation}
where $s_i(t)$ is the remaining-length score in Equation~\ref{eq:remaining}, $W_i(t)$ is accumulated queue waiting excluding execution, $\gamma=0.9$, and $\tau=30$\,s. Lower scores advance first.
The rule uses \method{TIE}'s aging formula and constants.
The \method{TIE} control uses its bounded expected length under this common rule; its native CVaR and dynamic-$\beta$ policy remains a separate reference.
\method{LTR} produces signed ordinal scores, which require calibration for use as token-length estimates. \method{LTR} maps them monotonically to token lengths using isotonic calibration on 360 existing LMSYS validation requests; no target lengths fit this adapter.
\method{TRAIL} uses the frozen 40-bin initial head, not its online prediction/filter chain. \method{SSJF} retains source-training normalization and author tail-512 tokenization; \method{EGTP} retains its author prefix-feature extractor and frozen source head. The simulation retains the disclosed method-specific checkpoint training budgets.
LEAP and RLS share the same initial source head; every eight completed requests update the latter using only that arm's completed labels. Different completion orders therefore induce different causal update sequences.
Each finite arrival stream is drained to completion. The table includes modeled execution costs, while cached initial scores are assumed available on arrival, so the replay isolates prediction utility; real serving costs are reported separately.

\section{Experimental Protocol and Supporting Results}
\label{app:experimental-protocol}

\subsection{Shared evaluation protocol}
\label{app:shared-evaluation}
Training uses the method-specific budgets in Appendix~\ref{app:baseline-reproduction}, with 360 validation and 1,000 test requests per model--dataset setting. Splits exclude normalized-prompt and conversation-parent overlap; the ShareGPT test set covers 983 parents. Test labels are first-rollout lengths capped at 2,048 tokens. Repeated rollouts provide additional labels for the same training request. Readout standardization is fitted on source training data and frozen; regularization minimizes source-validation token MAE over $\{10^{-4},10^{-3},10^{-2},10^{-1},1,10\}$.

Figure~\ref{fig:ood-bars} uses 1,000 LMSYS test requests. The Alpaca comparison uses 1,000 held-out tests after 1,000 separate feedback labels; MATH and Aya use 4,200-request predict-before-feedback streams. The TRAIL results in this comparison use its initial head.

The representative serving panel uses 500 held-out ShareGPT requests and follows Section~\ref{sec:similar-serving} with 20,592 KV blocks and no artificial step waits. Requests arrive at 2.64984 requests/s and share recorded output trajectories. A missing candidate score makes the entire unprotected decision fall back to \method{FCFS}. Per-length analyses use $L\leq256$, $257\leq L\leq1,024$, and $L>1,024$ for short, medium, and long requests, respectively. Serving points are single-run measurements unless stated otherwise.

\paragraph{Target-map control.}
\label{app:target-map}
Log, square-root, and identity heads share the block-30 features and validation-only regularization selection. Test requests are not resampled by length. Short/long subgroup counts are 196/312 on ShareGPT and 467/188 on LMSYS; medium requests remain in the overall result. Table~\ref{tab:single30-maps} equally averages metrics computed separately within each dataset and subgroup.

\subsection{Finite-arrival LMSYS serving control}
\label{app:lmsys-rps4200}

Figure~\ref{fig:lmsys-rps4200} uses the same frozen first 1,024 held-out LMSYS requests and 544,047 recorded output tokens per point, with seed~0 and offered rates of 20, 40, 60, 80, and 100 requests/s. All four methods are measured at all five rates. The ordered request identities, generation seeds, and token trajectories are shared; only arrival spacing changes with the offered rate.

All methods use 512 resident requests, a 64-request decode cap, 20,592 KV blocks of size~16, a 4,096-token batch limit, and a 0.8 GPU-memory fraction. All plotted predictors publish a frozen initial total-length estimate for common post-prefill ranking, without a per-token countdown. They share wall-clock Aging with $\gamma=0.9$ and a 30-second time scale. Protection takes priority; if any unprotected candidate lacks an accepted score, the unprotected choice falls back to \method{FCFS}.

\method{LEAP} uses the frozen block-30 square-root readout on CPU; FCFS has no prediction head. The source-trained \method{TIE} checkpoint supplies a deterministic bounded expectation with a 2,048-token cap. Its encoder was trained with 512-token inputs and executes with a 2,048-token input limit in this control. At every rate, the same adapter splits logical TIE batches into physical forwards of at most eight, preserving logical padding, output order, and the expectation calculation. \method{LTR} uses the source-validation isotonic mapping described in Appendix~\ref{app:matched-aging}, with physical forwards of at most eight. Both predictors execute on serving GPU~0.

Client metrics use the planned-arrival origin defined in Section~\ref{sec:evaluation}. Per-token latency is each request's end-to-end latency divided by its actual output-token count before taking the mean or P90; it is not inter-token latency. Throughput divides actual output tokens by the fully drained client duration. For the last three points, another job used GPU~1 and cores 12--15 while serving kept GPU~0 and cores 0--7.

\subsection{Input-width control}
\label{app:input-width}
The all-state input in Table~\ref{tab:input-width} follows Hugging Face's hidden-state convention: the embedding, raw decoder blocks 1--35, and the normalized final state of block 36. The same original labels, source-only standardization, six validation-selected regularization candidates, and capped round-0 test lengths are used for both inputs.

The input-ablation replay uses a single nonpreemptive server with service time equal to cached output length and initial scores available for free at arrival. A fixed seed (42) supplies common exponential spacings for offered loads $0.5,0.8,0.95,1.1$, plus simultaneous arrivals; 500 requests form a nested subset of the 1,000-request cohort. Prefill, batching, KV capacity, feature extraction, transfer, and prediction costs are excluded. For the 1,000-request all-state square-root comparison, P95 response changes range from a 10.36\% improvement to an 8.23\% increase.

Block 30 is the pre-existing default, not selected anew on these tests.

\subsection{Predictor Training Configurations}
\label{app:baseline-reproduction}

We use author implementations or their training cores, retaining the objectives and checkpoint-selection rules described below. These configurations apply to Table~\ref{tab:prediction-quality} and the LMSYS-source transfer results in Figure~\ref{fig:ood-bars}; serving and component controls use the checkpoints specified in their protocols. Optimizer configurations, implementation versions, and training traces are documented with the artifact.
\method{TIE} uses 20K LMSYS requests; \method{LTR} and \method{SSJF} use 10K ShareGPT/LMSYS requests per backbone. Other fits use 2,880 requests.

\paragraph{Readout configurations.}
The single-label Qwen scale control uses blocks 24, 24, 32, 30, 36, and 60 for the 0.6B, 1.7B, 4B, 8B, 14B, and 32B models, respectively. The default Qwen3-8B head has 4,096 inputs, with regularization 1.0 for ShareGPT and 0.1 for LMSYS; Mistral uses block 28. Whole-service resource controls retain their documented input configurations.

\paragraph{Text-based predictors.}
\method{TIE} trains its DeBERTa encoder and two distribution heads from pretrained initialization, following the author's encoder-freezing schedule. Its 20K LMSYS fits use 20 outputs per prompt. Point predictions use the bounded expectation with a 2,028-token cap; the LMSYS quality results use numerical integration. \method{LTR} uses 2,048-token inputs and ListMLE. \method{SSJF} uses the BERT REG-L1 recipe with tail-512 tokenization, training the encoder for three epochs and freezing it for three more. Our implementation standardizes targets using source-training statistics and applies the inverse transform at inference; quality results use un-clipped predictions. In the transfer replay, \method{LTR} converts ordinal scores to lengths by source-validation isotonic calibration, as specified in Appendix~\ref{app:matched-aging}.

\paragraph{SSJF target standardization.}
\label{app:ssjf-normalization}
An early diagnostic compared raw token-length targets with source-training target standardization using 2,880 training, 360 validation, and 1,000 test requests per setting.
Both variants retained the author's BERT REG-L1 network, tail-512 tokenization, six-epoch training loop, encoder freezing after epoch three, learning-rate schedule, and final-checkpoint rule, with seed 42.
Standardization used only the training-set mean and standard deviation; predictions were transformed back to token units before evaluation.
Standardization improved MAE and Spearman correlation in all five tested settings, motivating its use in our SSJF implementation; Table~\ref{tab:ssjf-normalization} shows the two Qwen3-8B settings, using un-clipped predictions against first-rollout test lengths.

\begin{table}[!htb]
\caption{SSJF target standardization at the early 2,880-request training budget.}
\label{tab:ssjf-normalization}
\centering
\small
\setlength{\tabcolsep}{2.5pt}
\renewcommand{\arraystretch}{1.08}
\begin{tabularx}{\linewidth}{@{}llCCCC@{}}
\toprule
& & \multicolumn{2}{c}{Raw targets} & \multicolumn{2}{c}{Standardized} \\
\cmidrule(lr){3-4}\cmidrule(lr){5-6}
Backbone & Dataset & MAE $\downarrow$ & $\rho\uparrow$ & MAE $\downarrow$ & $\rho\uparrow$ \\
\midrule
Qwen3-8B & ShareGPT & 779.39 & 0.0845 & 308.89 & 0.6716 \\
Qwen3-8B & LMSYS & 517.79 & 0.1658 & 240.90 & 0.7844 \\
\bottomrule
\end{tabularx}
\end{table}

No epochs, learning rates, clipping, or mappings were selected on test results.

\paragraph{Hidden-state predictors.}
The dynamic Qwen control uses block-30 trajectories and 360 validation requests per domain; EWMA gains of 0.01/0.05 for ShareGPT/LMSYS are selected on validation only. A separate eight-setting \method{TRAIL} diagnostic uses 1,000 test requests per setting, matched live token positions, and the same remaining-budget clip; it improves MAE over frozen \method{LEAP} countdown in six settings.

\method{TRAIL} includes a 40-bin initial MLP and an online head trained on intermediate generation states, with the original probability filter. Initial-head quality results remain distinct from evaluations of the complete prediction chain. \method{EGTP} follows the LP\_Bench implementation: entropy-weighted pooling of the first four raw prompt tokens, a two-layer 20-bin MLP, and the $0.95\,\mathrm{KL}+0.05\,\mathrm{MSE}$ objective. Its training batch sizes are 16 for Qwen/Llama and 256 for Mistral, with 10 training epochs on Qwen/Llama and 200 on Mistral.

\balance
\subsection{CPU pipeline measurement details}
\label{app:current-cpu-cost}
The component study assigns one prediction core per method, separate producer, consumer, and sampling cores, and one thread per numerical library. Each plotted cell offers batches of eight requests for ten seconds at rates of 1/10/100/1k/10k/100k per second, with eight queued batches and up to sixty seconds of drain. All paths reuse 64 source-validation prompts and predictors trained on 2,880 requests. Timing starts when raw text or the GPU representation is available and ends at consumer receipt; serving-model execution is excluded.

\method{LEAP} transfers a 4,096-coordinate FP32 feature to its FP64 affine routine. \method{TRAIL} transfers BF16 prompt states and pools them on CPU. Text predictors include tokenization and encoder execution; \method{TIE} uses 10,000 Monte Carlo draws for its capped expectation, excluding CVaR. Both \method{EGTP} variants use the author's four-token prefix: \method{EGTP-P} reconstructs logits on CPU, whereas \method{EGTP-L} assumes GPU-resident logits and hidden states. \method{EGTP-P}'s FP32 projection occupies 2.49\,GB beyond prediction-head storage. GPU pooling is evaluated as a separate feature-availability control.

Throughput counts receipts during the supply window; latency includes accepted batches completed during drain. Rejected offers, generator skips, and supply-end backlog remain in the accounting. CPU measurements separate prediction-process and consumer CPU time from producer and sampler work, and exclude initialization. Each cell is one hardware observation; generator-limited cells report delivered throughput without establishing predictor saturation.

The plotted seven pipelines were uniformly remeasured with the same process-isolated producer, receiver, and sampler. Offers are clocked by an integer native loop on the reserved producer core, using the same monotonic clock and arrival schedule. The producer continuously polls on its single reserved core at every offered rate, including the same one-second lead-in before each ten-second supply window. No rate uses a sleep branch. Admission telemetry is buffered during supply and written afterward; actual offer timestamps and skipped arrivals are retained. The same waiting policy applies across the entire offered-rate grid. The input FIFO retains eight queued batches and immediate queue-full rejection. Request transport and result publication use direct pipes, avoiding background queue-feeder threads; result serialization is charged to the same prediction process. The numerical worker, model weights, precision, inputs, offered-rate grid, ten-second supply, sixty-second drain, and absence of TTL remain unchanged. Original and intermediate measurements are preserved separately; none are mixed into this figure.

Generator qualification retains the original criteria: a cell is limited if more than 1\% of scheduled batches are skipped or if the P95 attempt lag exceeds one batch interval. Every newly plotted cell passes these criteria. Accepted work, rejected batches, generator skips, supply-end backlog, post-supply completions, all original timestamps, and all warnings remain in the artifact. Prediction workers and the separate receiver/sampler processes inherit the verified resource leases; CPU roles remain 12/13/14/15, with source representations on GPU1 and no serving-model execution.

\subsection{Prefill-overlap measurement}
\label{app:overlap-protocol}
Throughput characterizes CPU prediction capacity; overlap characterizes score availability. We measure the completion of CPU prediction and initial batch ordering relative to the remaining GPU prefill.
On Qwen3-8B, a native CPU worker reads block-30 features, computes scores and a stable batch order, and publishes a ready flag.
A GPU probe reads this flag immediately before the final RMSNorm, confirming CPU completion while mandatory prefill work is still pending.
Figure~\ref{fig:prefill-overlap-readiness} reports the counts separately for FP32 and FP16 asynchronous feature transport. Each mode and batch size has 24 executions across three temporal blocks in one A800 session, reusing fixed source-validation prompt groups. The FP32 control captures the same layer's features but defers transfer and CPU work until prefill returns; its zero readiness is the expected negative control.

\end{document}